\documentclass{article}
\usepackage{iclr2027_conference,times}

\usepackage{amsmath,amsfonts,bm}

\def\eqref#1{equation~\ref{#1}}

\def\1{\bm{1}}

\DeclareMathAlphabet{\mathsfit}{\encodingdefault}{\sfdefault}{m}{sl}
\SetMathAlphabet{\mathsfit}{bold}{\encodingdefault}{\sfdefault}{bx}{n}

\usepackage{url}
\usepackage{xcolor}
\usepackage{booktabs}
\usepackage{graphicx}
\usepackage{placeins}
\usepackage{float}
\usepackage{hyperref}

\title{UniTele: Learning Device-Specific Degradations for Unified Smartphone Telephoto Super-Resolution}

\author{%
Yuchen Zhang$^{1,2}$, Aojie Li$^{2,\dagger}$, Wei Cheng$^{2}$, Xianfang Zeng$^{2}$,\\
\bfseries Runxing Liu$^{2}$, Haixin Hao$^{3}$, Gang Yu$^{2,\ddagger}$, Xiaoyun Yuan$^{1,\ddagger}$\\[3pt]
\normalfont $^{1}$Shanghai Jiao Tong University\quad $^{2}$StepFun\\
\normalfont $^{3}$Communication University of Zhejiang
}

\iclrfinalcopy
\newcommand{\draftplaceholder}[1]{%
  \par\noindent\textcolor{gray}{\textit{Draft placeholder: #1}}\par}

\begin{document}

\maketitle
\lhead{}
\begingroup
\renewcommand{\thefootnote}{}
\footnotetext{$\dagger$ Aojie Li leads this project; $\ddagger$ Corresponding authors.}
\endgroup

\begin{abstract}
Super-telephoto imaging on mobile phones suffers from complex, device-dependent degradations that are poorly captured by conventional handcrafted pipelines. Collecting large-scale aligned training pairs across different phones and diverse scenes, however, is both costly and difficult to scale. We therefore study whether device-specific degradations can be learned from only a small number of physical captures and transferred for training a unified super-resolution (SR) model. Our key idea is to factor degradation learning into two learnable components: one captures the device-specific imaging characteristics under well-resolved conditions, while the other captures the additional information loss and artifacts that arise in super-telephoto imaging. We build a paired benchmark across three phone brands (Huawei, Apple, and vivo) by photographing the same content at two display sizes and zoom settings. The larger displayed image is captured at the phone's maximum optical zoom setting, while the smaller image is photographed directly at the phone's highest native zoom setting through the phone's native imaging pipeline. The benchmark contains 519 training and 60 held-out pairs. We use the benchmark to learn device-specific degradation, then apply it to external images to synthesize thousands of additional pairs to train our UniTele SR model. Relative to the pretrained Qwen and FLUX backbones before adaptation, the full UniTele training recipe substantially improves paired restoration fidelity and perceptual similarity on the held-out benchmark and yields markedly more stable and visually faithful restorations.
\end{abstract}

\section{Introduction}
\label{sec:introduction}
Recovering fine detail beyond the optical zoom range is a central challenge in smartphone super-telephoto imaging.  Generative super-resolution (SR) offers a promising approach by using learned image priors to reconstruct detail from severely degraded inputs~\citep{wang2023stablesr,wu2024seesr}. However, visually plausible detail is not necessarily faithful to the photographed scene: ambiguous observations can lead to incorrect textures or structures.  Adapting these models to smartphone telephoto restoration therefore calls for supervision that connects real degraded observations to their corresponding reference detail.

\begin{figure}[tp]
  \centering
  \includegraphics[width=\linewidth]{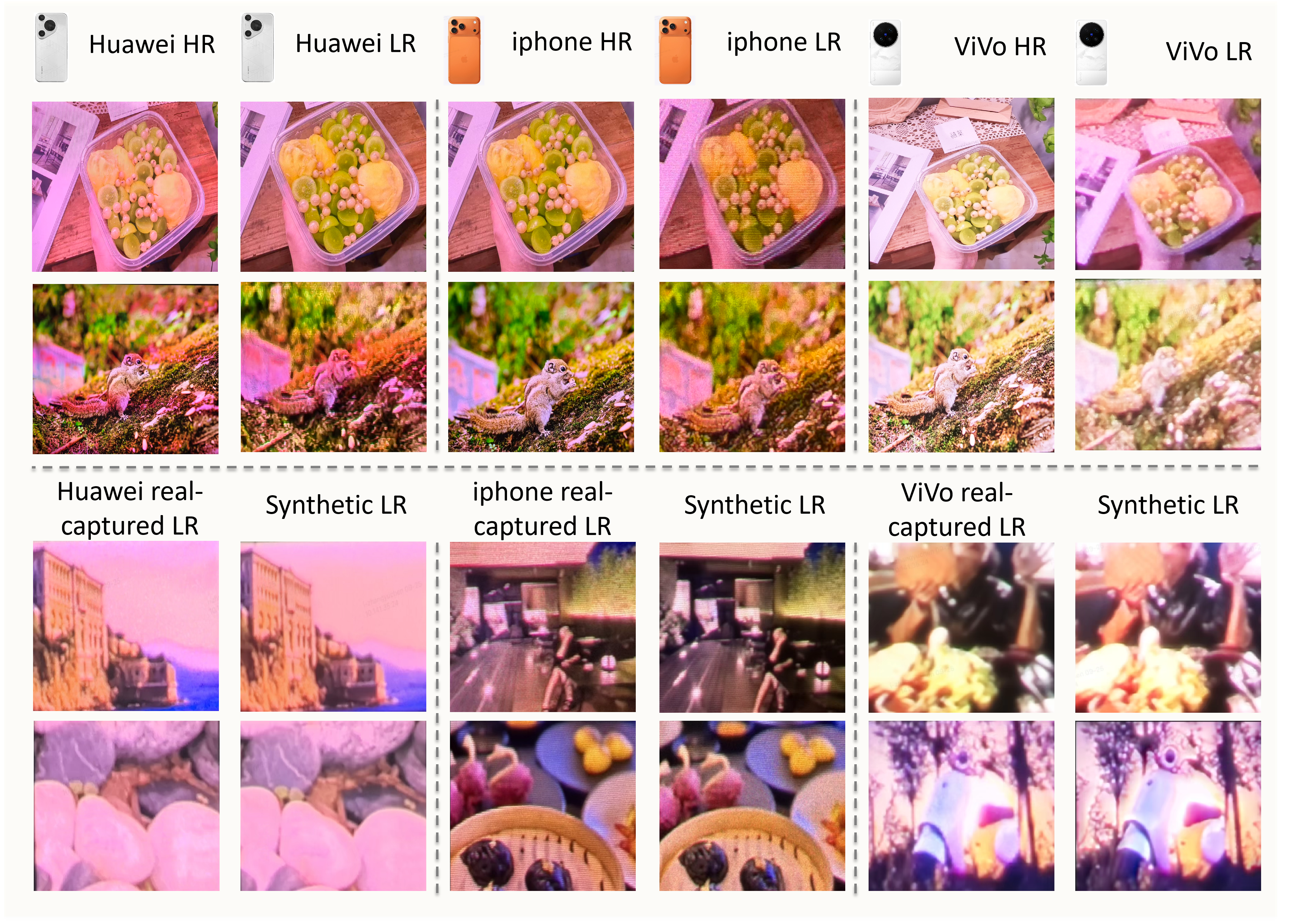}
  \caption{Paired HR (left) and LR (right) observations of the same content across Huawei, iPhone, and vivo phones.}
  \label{fig:device_vis}
\end{figure}

Obtaining such supervision is difficult because telephoto degradations are device dependent.  The same content can exhibit different color shifts, blur, and texture loss across phones (Figure~\ref{fig:device_vis}).  Paired captures provide direct examples of these differences, but collecting and aligning them across devices and diverse scenes is costly~\citep{cai2019realsr,wei2020drealsr,chen2019camerasr}. Handcrafted degradation pipelines readily generate pairs from large image collections, yet their predefined operators may not adequately capture the target device's imaging characteristics~\citep{zhang2021bsrgan,wang2021realesrgan}. Learned degradation models offer a way to synthesize more realistic training data~\citep{fritsche2019dsgan,wolf2021deflow}.  Building on this direction, we investigate how limited physical captures can support device-specific degradation synthesis on new content, extending the supervision available for generative telephoto restoration.

\begin{figure}[tp]
  \centering
  \includegraphics[width=\linewidth]{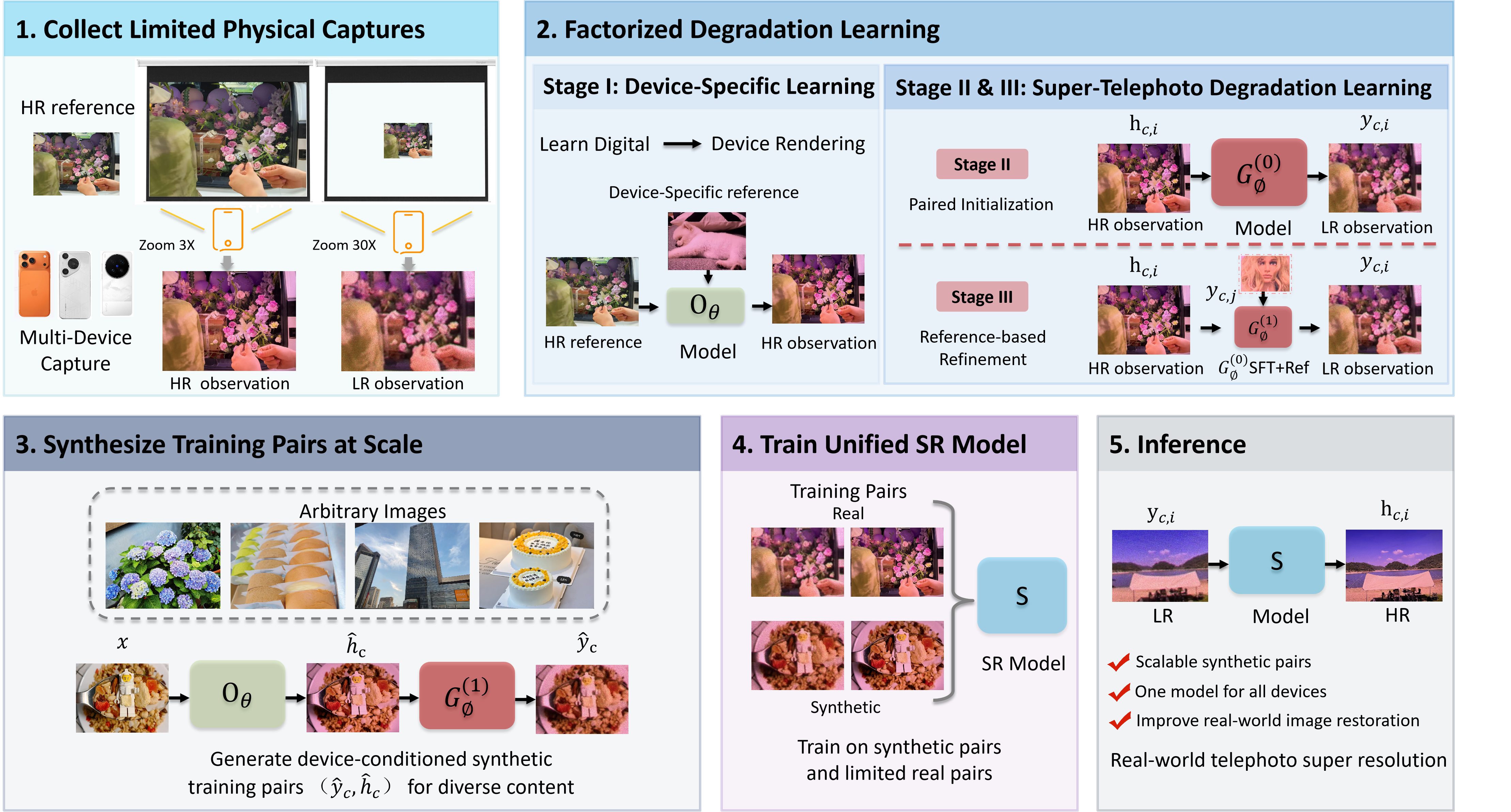}
  \caption{Overview of UniTele. Two learned mappings synthesize low-resolution (LR) images for SR training, using their corresponding synthetic high-resolution (HR) observations as targets.}
  \label{fig:method_overview}
\end{figure}

We propose UniTele, a framework that learns device-specific degradations from a small capture set and transfers them to external images for SR training.  Our key idea is to learn the device's imaging characteristics and the additional degradation associated with super-telephoto imaging separately. The first mapping learns the device's imaging characteristics under well-resolved conditions; the second learns the additional detail loss and artifacts in the corresponding super-telephoto input.

To learn these mappings, we construct a controlled paired benchmark using Huawei, Apple, and vivo phones.  We denote the original image displayed on the monitor as the \emph{high-resolution reference}.  Each image is photographed at two display sizes and zoom settings. The larger image is captured at the phone's maximum optical zoom setting to obtain the \emph{high-resolution observation}; the smaller image is captured directly at the phone's highest native zoom setting to obtain the \emph{low-resolution observation}. The benchmark contains 519 training pairs and 60 held-out pairs. The first mapping is learned from high-resolution references to high-resolution observations, and the second from high-resolution observations to low-resolution observations.  We then apply the learned mappings to external images to synthesize thousands of additional pairs. These synthetic pairs are combined with real pairs to adapt pretrained generative models to telephoto SR.

Experiments on the held-out paired benchmark demonstrate improved restoration fidelity and perceptual similarity across both Qwen and FLUX backbones.  Qualitative comparisons further illustrate improvements in the preservation of scene structure.

Our contributions are threefold:
\begin{itemize}
  \item We propose a device-conditioned degradation framework that separately learns device-specific imaging characteristics and super-telephoto degradation from limited captures.
  \item We construct a controlled smartphone super-telephoto benchmark spanning Huawei, Apple, and vivo to characterize device-dependent degradations and evaluate restoration across different phones.
  \item We demonstrate the effectiveness of UniTele on Qwen and FLUX backbones, with improvements in reconstruction fidelity, perceptual similarity, and visual quality.
\end{itemize}

\section{Related Work}
\label{sec:related_work}

\paragraph{Real-world super-resolution and optical zoom.}
Classical single-image super-resolution (SISR) creates low-resolution (LR) observations by applying spatially invariant blur and bicubic downsampling to high-resolution (HR) images. This controlled formulation omits coupled effects of optics, sensor noise, image signal processing, sharpening, compression, and spatially varying blur. RealSR instead captures the same scene at different focal lengths and progressively registers the pairs for supervision~\citep{cai2019realsr}. DRealSR broadens real degradation diversity and shows that smooth regions, edges, and textures pose different restoration challenges~\citep{wei2020drealsr}. Camera Lens Super-Resolution (CameraSR) studies the lens resolution--field-of-view trade-off. It shows that realistic lens pairs better support digital zoom than conventional synthetic downsampling~\citep{chen2019camerasr}. Related mobile systems align and adaptively fuse concurrent wide and telephoto captures to improve intermediate zoom~\citep{wu2024hybridzoom}. These studies establish the importance of real optical observations and paired acquisition. Our task differs from fixed-scale SISR: aligned inputs and targets may share a canvas but differ in effective optical detail and camera processing. We use controlled, calibration-assisted captures to learn a forward model that transfers measured device characteristics to digital source content.

\paragraph{Synthetic and learned degradation modeling.}
Practical blind SR methods attempt to close the synthetic-to-real gap by enriching the forward degradation process. BSRGAN combines blur, resampling, sensor-inspired noise, and JPEG compression in randomly shuffled orders~\citep{zhang2021bsrgan}, while Real-ESRGAN introduces a high-order degradation pipeline and explicitly models ringing and overshoot artifacts~\citep{wang2021realesrgan}. Such pipelines are efficient and scalable, but their realism is bounded by a predefined operator family and manually chosen parameter ranges. Learned degradation models relax these assumptions. DSGAN uses frequency separation and adversarial learning to transfer real-image characteristics to synthetically downsampled images while preserving low-frequency content~\citep{fritsche2019dsgan}. DeFlow instead models stochastic conditional degradations in a shared latent space and learns from unpaired clean and degraded marginals using conditional normalizing flows~\citep{wolf2021deflow}. These approaches typically learn one aggregate target degradation distribution. In contrast, mobile telephoto degradations are not interchangeable: different phones, lenses, zoom levels, and ISP pipelines can produce distinct tone, blur, sharpening, compression, and residual texture for the same scene. We therefore study a single conditional generator whose output distribution is indexed by device identity and anchored by a telephoto degradation reference from that device. This formulation enables explicit device switching without maintaining one degradation model per phone.

\paragraph{Generative priors and conditional image restoration.}
Large pretrained generative models have recently become strong priors for real-world image restoration. StableSR adapts a pretrained text-to-image diffusion model to blind SR while introducing controllable feature wrapping to balance perceptual quality and fidelity~\citep{wang2023stablesr}. SeeSR extracts degradation-aware semantic prompts to reduce semantic errors under severe input degradation~\citep{wu2024seesr}, and the One-Step Effective Diffusion Network (OSEDiff) distills the restoration process into a single diffusion step for efficient inference~\citep{wu2024osediff}.  Adversarial Diffusion Compression for Real-World Image Super-Resolution (AdcSR) further compresses one-step diffusion through structural pruning and adversarial distillation~\citep{chen2025adcsr}. In parallel, all-in-one restoration methods use learned prompts or language instructions to select among restoration behaviors. PromptIR learns degradation-specific prompts for a unified restoration network~\citep{potlapalli2023promptir}, whereas InstructIR conditions restoration directly on human-written instructions~\citep{conde2024instructir}. These methods primarily use generative or conditional priors in the inverse direction, mapping an unknown degraded image to a clean output. Our use of an image-editing foundation model, instantiated with Qwen-Image-Edit-2511~\citep{wu2025qwenimage,qwen2025imageedit2511}, is complementary: we adapt the prior in the forward direction to map digital source content to a phone-specific synthetic LR image.

\section{Method}
\label{sec:method}
\label{sec:benchmark_task}
We train a unified SR model using limited paired captures and external images without phone observations. UniTele learns device-specific degradations from the captures and applies them to external content to expand paired supervision. Figure~\ref{fig:method_overview} shows controlled acquisition, two learned mappings for pair synthesis, and joint training on real and synthetic pairs.

\subsection{Multi-Device Paired Acquisition}
\label{sec:data}
\label{sec:capture}
\label{sec:alignment}
\label{sec:benchmark_characterization}
\label{sec:benchmark_splits}
Learning device-specific degradations requires paired observations with reliable content correspondence, but obtaining a well-resolved counterpart to an in-the-wild super-telephoto image is difficult. We therefore vary the size of an image displayed on a monitor to simulate the apparent size of the same subject at different distances: a smaller image represents a more distant subject, while a larger image provides a closer view of the same content.

Let $x$ denote the original digital image, or \emph{high-resolution (HR) reference}. Keeping the camera--monitor distance fixed, we photograph the larger displayed image with device $c$ at its maximum optical zoom setting to obtain the \emph{HR observation} $h_c$. We then photograph the smaller displayed image directly at the phone's highest native zoom setting to obtain the \emph{low-resolution (LR) observation} $y_c$. The displayed image sizes are chosen so that the subject occupies roughly the same fraction of the frame in $h_c$ and $y_c$. In this pair, $y_c$ represents a distant subject photographed at the phone's highest native zoom, while $h_c$ provides a well-resolved observation of the same content. We photograph a calibration board at both zoom settings and estimate a homography from corresponding corners to align $h_c$ and $y_c$. The resulting $(x,h_c)$ pairs provide supervision for the device's imaging characteristics under well-resolved conditions, while $(h_c,y_c)$ pairs supervise the additional detail loss and artifacts associated with super-telephoto imaging.

The benchmark includes 519 training pairs and 60 test pairs from Huawei Pura 70 Pro+, iPhone 17 Pro Max, and vivo X300 Ultra, with 20 test pairs per device.

\subsection{Factorized Degradation Learning}
\label{sec:degradation_method}
\label{sec:formulation}
The goal of factorized degradation learning is to synthesize device-specific HR and LR observations from an HR reference. We formulate this process as two mappings:
\begin{equation}
  \hat h_c=O_{\theta}(x,c,r_c^{\mathrm{HR}}),\qquad
  \hat y_c=G_{\phi}(\hat h_c,c,r_c^{\mathrm{LR}}).
  \label{eq:method_chain}
\end{equation}
Here $\hat h_c$ and $\hat y_c$ are synthetic HR and LR observations. For each device, a fixed HR--LR image pair $(r_c^{\mathrm{HR}},r_c^{\mathrm{LR}})$ provides references for the device-specific imaging style and super-telephoto degradation style, respectively. The two mappings are trained in three stages (Figure~\ref{fig:method_overview}).

\label{sec:optical_mapping}
\label{sec:stage1}
\label{sec:stage2}
\label{sec:stage3}
Stage I (Device-Specific Learning) learns the phone's color response and texture rendering by training $O_{\theta}$ on $x\rightarrow h_c$ pairs, with $x$ and $r_c^{\mathrm{HR}}$ as image inputs. Here $c$ denotes the device name (Huawei, Apple, or vivo), which is provided through Qwen's text prompt. The same prompt instructs the model to preserve the scene content, geometry, and text in $x$. Stage II (Paired Initialization) establishes the HR-to-LR mapping for super-telephoto detail loss and artifacts: we initialize $G_{\phi}$ from the pretrained model and train it on $(h_c,c)\rightarrow y_c$ pairs without an LR style reference. Stage III (Reference-based Refinement) adds direct visual conditioning for the phone's degradation style by continuing to train the same $G_{\phi}$ adapter with $r_c^{\mathrm{LR}}$ as an additional image input, retaining the paired supervision.

We train one LoRA adapter for $O_{\theta}$ and another for $G_{\phi}$ on Qwen-Image-Edit-2511~\citep{qwen2025imageedit2511}, keeping the backbone weights frozen and using the standard conditional flow-matching objective. Each adapter is shared across devices, with balanced device sampling during training. Evaluation and synthesis use a fixed seed.

\subsection{Super-Resolution Training with Real and Synthetic Data}
\label{sec:sr_method}
\label{sec:synthetic_data}
To expand SR supervision without additional captures, we apply Equation~\ref{eq:method_chain} to external HR references $x$, generating synthetic training pairs $(\hat y_c,\hat h_c)$ with the corresponding synthetic HR observation as the restoration target. These pairs supplement captured pairs $(y_c,h_c)$, broadening content coverage while retaining real degradation examples. We use pretrained Qwen and FLUX models for SR reconstruction, adapting each through LoRA to recover the HR observation from its LR observation. Each model is first trained on captured pairs and then further trained on the combined data across the target devices. The Stage I ablation in Table~\ref{tab:method_ablations} supports the two-step synthesis design: using $G_{\phi}$ alone to map $x$ directly to $\hat y_c$ yields substantially lower synthesis fidelity.

\section{Experiments}
\label{sec:experiments}
We examine real--synthetic training across SR backbones, in-the-wild transfer, the forward-model stages, and cross-device generalization.

\subsection{Experimental Setup}
\label{sec:experimental_setup}
\label{sec:sr_backbones}
\label{sec:metrics}
Experiments use 519 captured training pairs, 60 paired test images (Section~\ref{sec:data}), and a separate in-the-wild set. We evaluate Qwen-Image-Edit-2511 (Qwen SR)~\citep{qwen2025imageedit2511} and FLUX.2 [klein] 4B Base (FLUX.2 SR)~\citep{bfl2026flux2klein}. We compare pretrained backbones with LoRA adapters trained on real, synthetic, or mixed pairs; the mixed models are UniTele (Qwen) and UniTele (FLUX). The Real-ESRGAN baseline trains Qwen SR on pairs synthesized with its degradation pipeline~\citep{wang2021realesrgan}. Official pretrained AdcSR~\citep{chen2025adcsr} and OSEDiff~\citep{wu2024osediff} provide further comparisons.

\noindent\textbf{Qwen SR training.}
We adapt Qwen using conditional flow-matching with timestep-weighted MSE, applying LoRA to the attention Q/K/V and output projections, added Q/K/V and output projections, image and text MLP output projections, and modulation layers with rank 32, scaling factor $\alpha$ 32, and dropout 0. Training uses AdamW with learning rate $10^{-4}$ in the captured-data phase and $5\times10^{-5}$ in the mixed-data phase, a constant schedule, an effective batch size of 8, and image resolution set dynamically up to $786{,}432$ pixels (typically $1024\times768$ in landscape or $768\times1024$ in portrait). The captured-data and mixed-data phases run for 3,055 and 2,340 updates, respectively. Training uses 8 Ascend 910B2C NPUs and takes approximately 4.08 hours for the first phase and 3.13 hours for the second.

\noindent\textbf{FLUX SR training.}
We adapt FLUX using conditional flow-matching with timestep-weighted MSE, applying LoRA to dual-stream attention projections, added Q/K/V projections, input and output projections, joint attention--MLP projections, and the attention output projections in 20 single-transformer blocks with rank 32, scaling factor $\alpha$ 32, and dropout 0. Training uses AdamW with learning rate $10^{-4}$, a constant schedule, an effective batch size of 8, and image resolution set dynamically up to $262{,}144$ pixels (typically about $576\times432$). The captured-data and mixed-data phases run for 2,466 and 2,531 updates, respectively. Training uses 8 Ascend 910B2C NPUs and takes approximately 0.57 hours for the first phase and 0.55 hours for the second.

For full-reference evaluation, we report RGB PSNR, SSIM, MAE, LPIPS~\citep{zhang2018lpips}, and the predicted-to-target high-frequency (HF) energy ratio. We report per-device scores and their arithmetic mean across devices; an HF ratio closer to one indicates better agreement in high-frequency energy. For Human Preference (HP), data-evaluation experts compare each output with its aligned GT and assign a 1--10 score based on visual similarity, content consistency, and color shift; we report the mean score. For in-the-wild images without paired targets, we report MUSIQ~\citep{ke2021musiq}, CLIP-IQA~\citep{wang2023clipiqa}, and NIQE~\citep{mittal2013niqe}, alongside qualitative comparisons.

\subsection{Downstream Super-Resolution Evaluation}
\label{sec:sr_results}

\begin{table}[!htb]
\centering
\scriptsize
\caption{SR performance on 60 paired test images. Ratios denote real-to-synthetic sampling; dashes indicate unavailable HP scores.}
\label{tab:sr_data_comparison}
\setlength{\tabcolsep}{2pt}
\begin{tabular}{llrrrrrr}
\toprule
Backbone & Training setting & PSNR $\uparrow$ & SSIM $\uparrow$ & MAE $\downarrow$ & LPIPS $\downarrow$ & HF ratio $\to1$ & HP $\uparrow$ \\
\midrule
Input & No restoration & 16.99 & 0.5234 & 27.10 & 0.5704 & 0.680 & -- \\
AdcSR~\citep{chen2025adcsr} & Official pretrained model & 16.90 & 0.5673 & 27.03 & 0.4383 & 0.698 & 2.623 \\
OSEDiff~\citep{wu2024osediff} & Official pretrained model & 16.92 & 0.5810 & 26.90 & 0.4617 & 0.641 & -- \\
\midrule
Qwen SR & Base (untrained) & 11.66 & 0.4931 & 54.16 & 0.6618 & 0.471 & 1.459 \\
 & Real-ESRGAN degradations & 16.90 & 0.5297 & 27.39 & 0.5288 & 0.740 & 2.262 \\
 & Real-world only & 18.84 & 0.6098 & 20.43 & 0.3187 & 0.740 & 5.180 \\
 & Learned synthetic only & 18.05 & 0.5976 & 22.05 & 0.3438 & 0.684 & 4.689 \\
 & UniTele (Qwen), $1{:}2$ & \textbf{19.33} & \textbf{0.6186} & \textbf{18.58} & \textbf{0.3078} & 0.729 & 5.689 \\
 & UniTele (Qwen), $2{:}1$ & 19.32 & 0.6174 & 18.81 & 0.3080 & 0.749 & 5.656 \\
\midrule
FLUX.2 SR & Base (untrained) & 10.59 & 0.3638 & 62.55 & 0.6102 & 1.573 & 3.475 \\
 & Real-world only & 16.53 & 0.5643 & 26.84 & 0.4254 & 0.761 & 5.951 \\
 & Learned synthetic only & 17.56 & 0.5810 & 23.73 & 0.3958 & 0.690 & 5.623 \\
 & UniTele (FLUX), $1{:}2$ & 17.56 & 0.6041 & 24.91 & 0.3735 & 0.615 & 6.066 \\
 & UniTele (FLUX), $2{:}1$ & 18.40 & 0.6103 & 21.54 & 0.3590 & 0.657 & 6.295 \\
\bottomrule
\end{tabular}
\end{table}

\begin{figure}[!htb]
  \centering
  \includegraphics[width=0.97\linewidth]{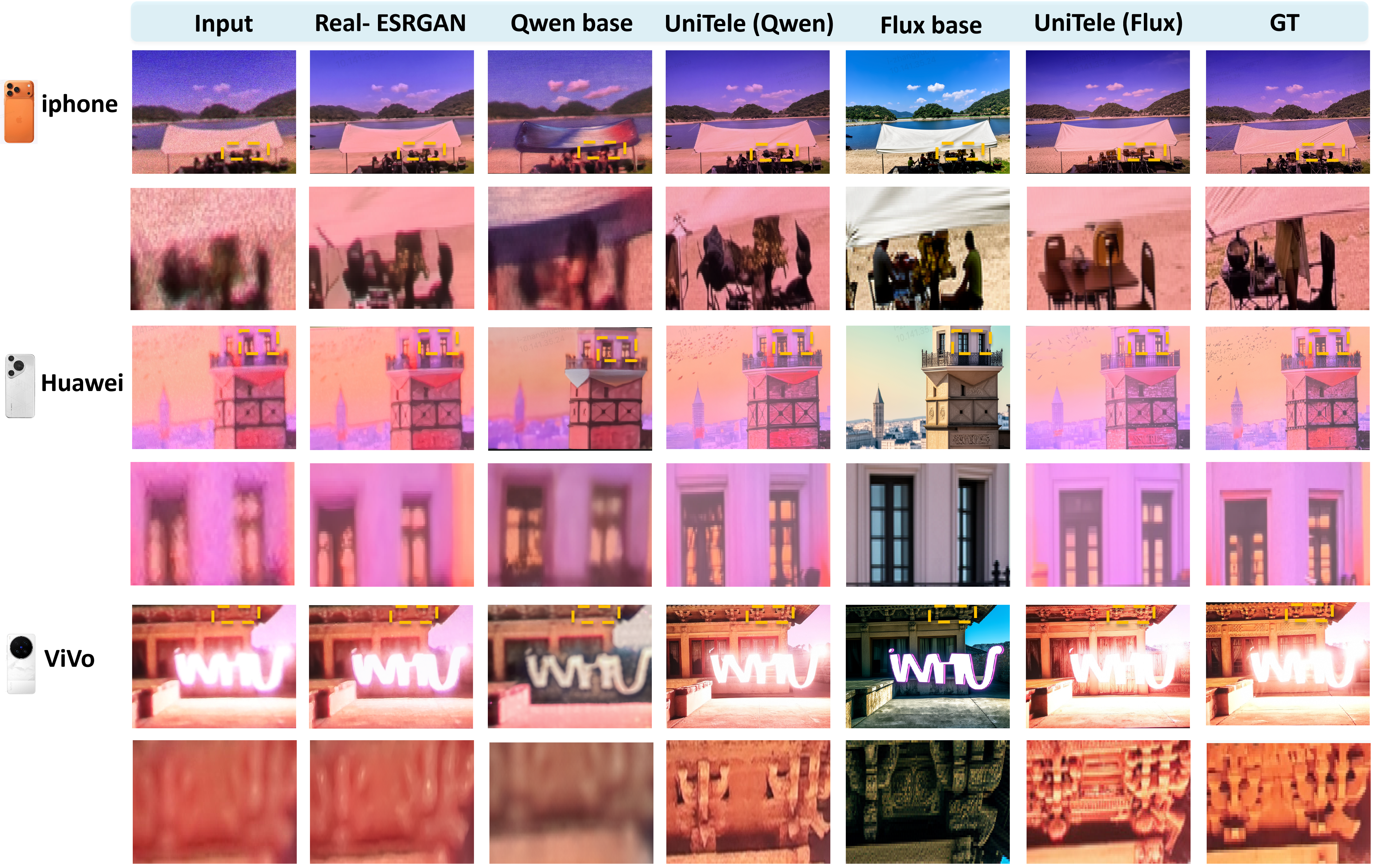}
  \caption{Qualitative SR comparison on the three-phone paired test set. GT denotes the aligned real-world HR image.}
  \label{fig:main_results}
\end{figure}

\begin{figure}[!htb]
  \centering
  \IfFileExists{compare.png}{%
    \includegraphics[width=0.8\linewidth]{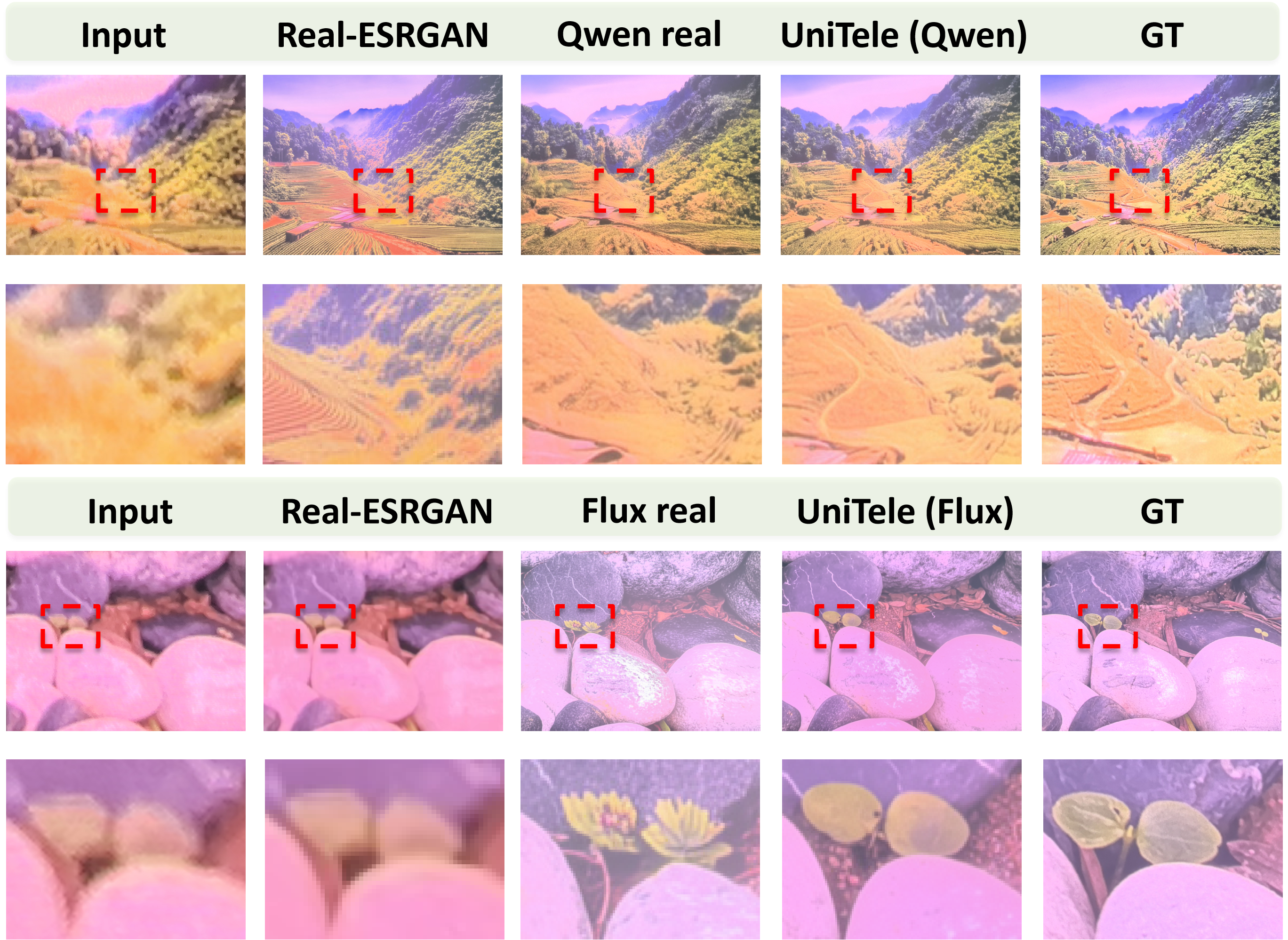}%
  }{%
    \fbox{\parbox[c][0.22\textheight][c]{0.88\linewidth}{\centering
      Reserved for the two-row training-ratio comparison.\\[0.5em]
      Add \texttt{main\_ratios.png} to replace this box.}}%
  }
  \caption{Effect of the real-to-synthetic sampling ratio on UniTele (Qwen) and UniTele (FLUX).}
  \label{fig:ratio_results}
\end{figure}

Table~\ref{tab:sr_data_comparison} compares each untrained base with real-world-only and synthetic-only training and two real-world--synthetic sampling ratios. The Qwen block additionally includes the Real-ESRGAN degradation-data baseline, while pretrained AdcSR and OSEDiff provide separate SR-model comparators.  UniTele (Qwen) at $1{:}2$ obtains the best PSNR, SSIM, MAE, and LPIPS: relative to real-world-only training, PSNR rises by 0.49~dB and LPIPS falls by 0.0109.  UniTele (FLUX) at $2{:}1$ also improves over its real-world-only counterpart.  These comparisons reflect complete training recipes. Figure~\ref{fig:main_results} places the input, selected model outputs, and real-world HR image side by side, exposing both recovered details and failure modes such as altered scene structure.

\subsection{Generalization to In-the-Wild Images}
\label{sec:unpaired_generalization}
We evaluate transfer from screen-based training to in-the-wild scenes on 16 iPhone LR images without paired HR targets, using fixed checkpoints without further adaptation. Both UniTele variants use a real-to-synthetic sampling ratio of $1{:}2$; GPT Image 2~\citep{openai2026gptimage2} provides an additional comparison. We report mean no-reference scores at $1024\times768$, resizing FLUX.2 outputs from their native working resolution of $576\times432$ before evaluation.

\begin{figure}[!htb]
  \centering
  \includegraphics[width=\textwidth]{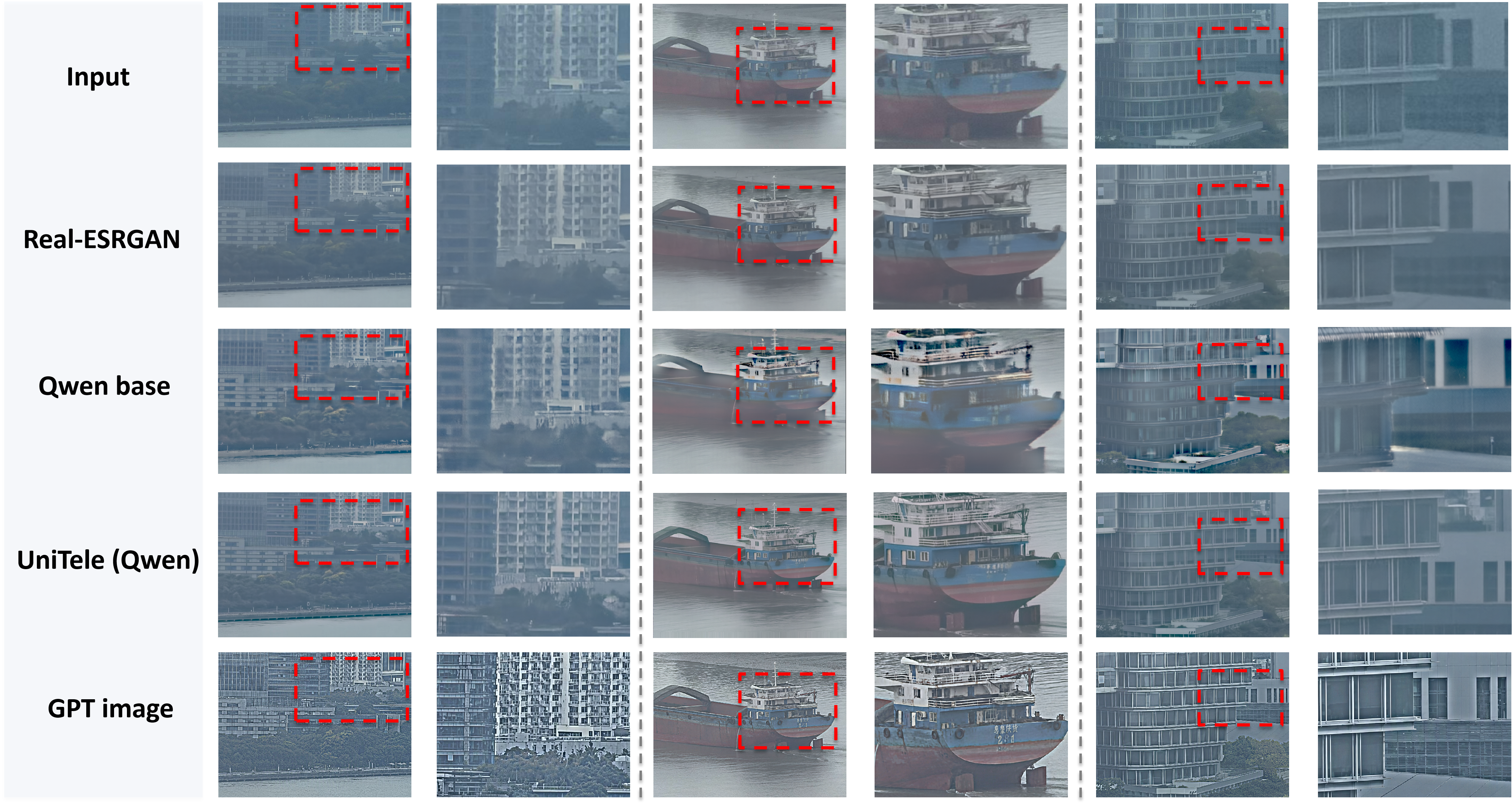}
  \caption{Qualitative SR comparison on in-the-wild images. Red boxes mark the regions shown in the adjacent close-ups.}
  \label{fig:unpaired_outdoor}
\end{figure}

\begin{table}[!htb]
\centering
\footnotesize
\caption{No-reference quality on 16 in-the-wild iPhone images.}
\label{tab:unpaired_sr}
\setlength{\tabcolsep}{2pt}
\begin{tabular}{lrrr}
\toprule
Method & MUSIQ $\uparrow$ & CLIP-IQA $\uparrow$ & NIQE $\downarrow$ \\
\midrule
Qwen SR base & 44.33 & 0.2101 & 6.89 \\
FLUX.2 SR base & 65.60 & 0.5569 & 5.35 \\
UniTele (Qwen), $1{:}2$ & 54.49 & 0.2335 & 7.14 \\
UniTele (FLUX), $1{:}2$ & 58.27 & 0.3721 & 5.85 \\
Qwen SR, Real-ESRGAN degradations & 45.88 & 0.1767 & 8.03 \\
GPT Image 2 & 67.83 & 0.5686 & 3.73 \\
\bottomrule
\end{tabular}
\end{table}

Table~\ref{tab:unpaired_sr} shows that UniTele (Qwen) improves MUSIQ and CLIP-IQA over its pretrained backbone, with a slight increase in NIQE. UniTele (FLUX) scores worse than its backbone on all three metrics, while GPT Image 2 obtains the best scores. In Figure~\ref{fig:unpaired_outdoor}, UniTele (Qwen) produces clearer window grids and ship railings than the Real-ESRGAN-trained baseline, with fewer conspicuous structural distortions than the Qwen base model. These examples illustrate transfer to real scenes, but no-reference scores and visual comparisons cannot verify the fidelity of recovered details without HR targets.

\begin{figure}[!htb]
  \centering
  \includegraphics[width=\textwidth]{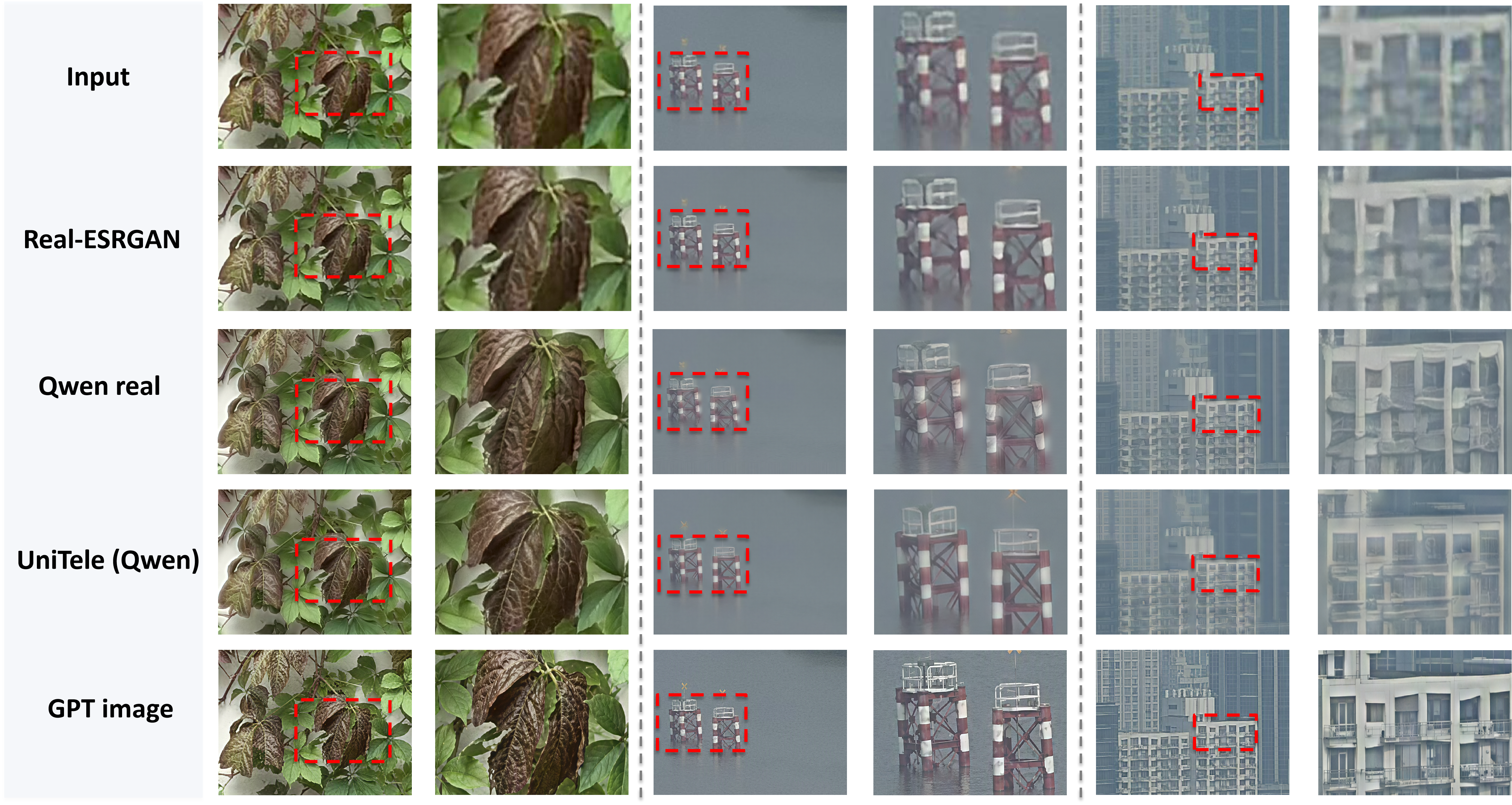}
  \caption{Additional in-the-wild SR comparisons. Rows show the input, Real-ESRGAN-trained Qwen, real-only Qwen, UniTele (Qwen), and GPT Image 2; red boxes mark the adjacent close-ups.}
  \label{fig:unpaired_vis2}
\end{figure}

Figure~\ref{fig:unpaired_vis2} provides additional matched crops for comparing local detail and structural consistency across the real-only and mixed-data Qwen models, alongside the other in-the-wild baselines.

\subsection{Ablation Studies}
\label{sec:ablations}
\paragraph{Training-device ablation.}
Using the Qwen SR backbone, we compare all-device training with training restricted to Vivo data.  We apply this restriction to both real-world-only training and UniTele (Qwen) at $1{:}2$, and evaluate every model on the complete three-device paired test set.  Initialization and training settings are matched, with the smaller single-device pool sampled repeatedly.  The UniTele comparison uses a separate controlled run from the main results.

\begin{table}[htbp]
\centering
\footnotesize
\caption{Effect of single- versus multi-device training on Qwen SR, evaluated across all three phones.}
\label{tab:cross_device_generalization}
\setlength{\tabcolsep}{2pt}
\begin{tabular}{llrrrrr}
\toprule
Device pool & Training data & PSNR $\uparrow$ & SSIM $\uparrow$ & MAE $\downarrow$ & LPIPS $\downarrow$ & HF ratio $\to1$ \\
\midrule
All three devices & Real-world only & 18.84 & 0.6098 & 20.43 & 0.3187 & 0.740 \\
Vivo only & Real-world only & 14.47 & 0.5451 & 38.99 & 0.4192 & 0.832 \\
\midrule
All three devices & UniTele (Qwen), $1{:}2$ & 18.81 & 0.6031 & 19.87 & 0.3199 & 0.766 \\
Vivo only & UniTele (Qwen), $1{:}2$ & 13.51 & 0.5363 & 45.06 & 0.4617 & 0.815 \\
\bottomrule
\end{tabular}
\end{table}

Restricting training to Vivo alone reduces PSNR by 4.37~dB for real-world-only training and 5.30~dB for UniTele (Qwen) on the three-device test set; LPIPS also rises by 0.1005 and 0.1418, respectively.  These large drops indicate limited generalization from single-device training to the multi-device benchmark, even when synthetic pairs are included.

\paragraph{Stage ablation.}
We evaluate the forward model directly, rather than only through downstream SR.  Removing Stage I feeds digital images directly to the telephoto generator; removing Stage III uses the reference-free Stage II model.  Both variants and the complete pipeline are evaluated against 50 source-matched real-world LR images from Huawei, iPhone, and vivo, using the same inference seed.

\begin{table}[htbp]
\centering
\footnotesize
\caption{Stage ablations for LR synthesis on 50 paired test images from three phones.}
\label{tab:method_ablations}
\setlength{\tabcolsep}{2pt}
\begin{tabular}{lrrrrr}
\toprule
Forward-model variant & PSNR $\uparrow$ & SSIM $\uparrow$ & MAE $\downarrow$ & LPIPS $\downarrow$ & HF ratio $\to1$ \\
\midrule
Without Stage I: direct digital input & 12.05 & 0.4657 & 54.39 & 0.6337 & 1.992 \\
Without Stage III: no reference-based refinement & 15.20 & \textbf{0.5473} & 36.05 & 0.4001 & 0.950 \\
Full three-stage model & \textbf{15.52} & 0.5439 & \textbf{35.01} & \textbf{0.3942} & \textbf{1.018} \\
\bottomrule
\end{tabular}
\end{table}

Bypassing Stage I causes the largest loss, reducing PSNR by 3.47~dB and increasing LPIPS by 0.2395 relative to the full pipeline.  Removing Stage III also reduces PSNR and increases LPIPS.  The full model has the best PSNR and LPIPS and the HF ratio closest to one, although the Stage-II-only variant has slightly higher SSIM.

\begin{figure}[!htb]
  \centering
  \includegraphics[width=\linewidth]{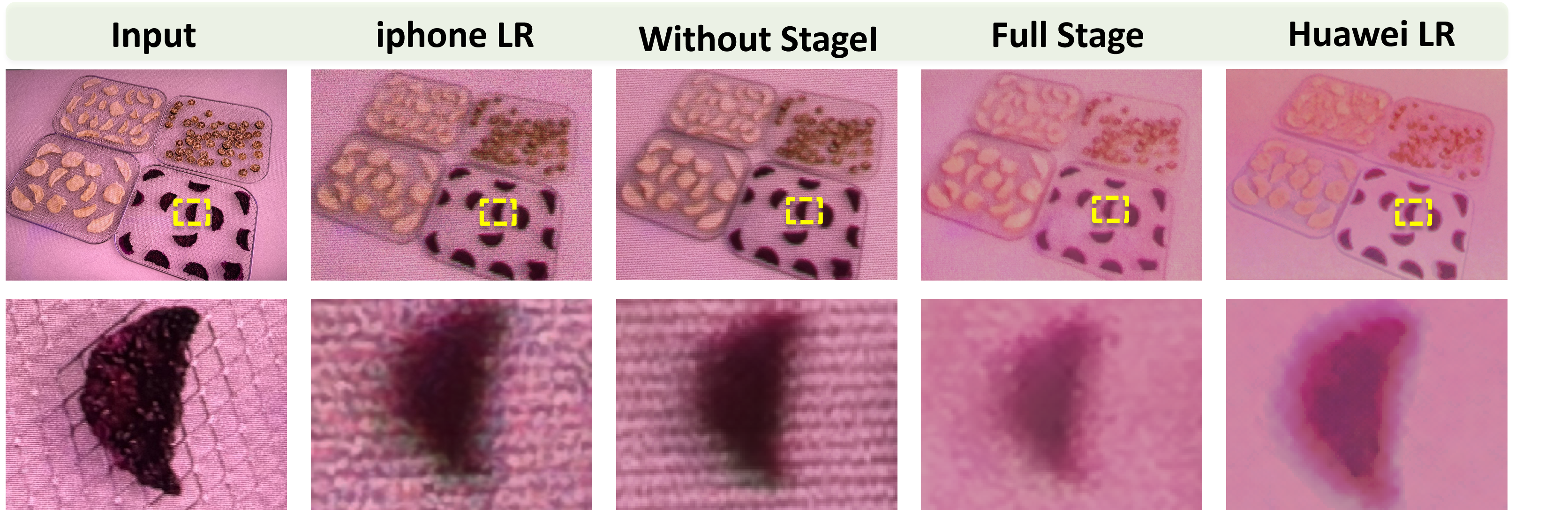}
  \caption{Stage I ablation for Huawei-targeted degradation synthesis. Columns show the digital input, iPhone LR, synthesis without Stage I, the full model, and Huawei LR; the bottom row shows yellow-box close-ups.}
  \label{fig:stage1_vis}
\end{figure}

Figure~\ref{fig:stage1_vis} illustrates the Stage I ablation on shared source content: the full pipeline more closely reproduces the Huawei observation's color and degradation appearance than direct digital input to the telephoto generator.

\section{Conclusion}
\label{sec:conclusion}
We presented UniTele, a device-conditioned approach that learns camera-domain rendering and super-telephoto degradation from a limited set of aligned captures, then transfers both transformations to external images to expand SR supervision. On the 60-pair three-phone test set, combining real and synthetic pairs improved PSNR and LPIPS over real-world-only training for both Qwen and FLUX backbones. Forward-model ablations support the factorized synthesis design, while the single-device ablation shows the value of training across phones for this benchmark. Results on in-the-wild images suggest qualitative transfer, although the no-reference scores are mixed and cannot establish whether recovered details match the scene. Together, these findings indicate that learned, device-conditioned pair synthesis can extend the utility of a small capture set for smartphone telephoto restoration.

\paragraph{Limitations.}
\label{sec:limitations}
The capture protocol is restricted to a planar display, and residual alignment error and display-specific artifacts may affect both supervision and evaluation. Device coverage is finite, and a generative degradation model may alter content.

\clearpage
\subsection*{AI use statement}
Generative AI models were used to synthesize device-conditioned training pairs, as described in Section~\ref{sec:method}. An AI writing assistant also assisted with manuscript editing, consistency and citation checks, and LaTeX formatting. The authors are responsible for verifying AI-assisted text, references, experimental claims, and generated artifacts against the underlying sources and results, and take responsibility for the final content of this paper.

\bibliography{references}
\bibliographystyle{iclr2027_conference}
\end{document}